\documentclass[sigconf]{acmart}
\usepackage{balance}
\AtBeginDocument{%
  \providecommand\BibTeX{{%
    \normalfont B\kern-0.5em{\scshape i\kern-0.25em b}\kern-0.8em\TeX}}}

\copyrightyear{2020}
\acmYear{2020}
\setcopyright{acmlicensed}\acmConference[MM '20]{Proceedings of the 28th ACM International Conference on Multimedia}{October 12--16, 2020}{Seattle, WA, USA}
\acmBooktitle{Proceedings of the 28th ACM International Conference on Multimedia (MM '20), October 12--16, 2020, Seattle, WA, USA}
\acmPrice{15.00}
\acmDOI{10.1145/3394171.3413778}
\acmISBN{978-1-4503-7988-5/20/10}
\newcommand{\model}{\mbox{\textsc{LIGHTEN}}}
\begin{document}
\fancyhead{}

\author{Sai Praneeth Reddy Sunkesula}
\affiliation{%
 \institution{Indian Institute of Technology Bombay}
 \city{Mumbai}
 \state{Maharashtra}
 \country{India}}
\email{praneeth20@cse.iitb.ac.in}

\author{Rishabh Dabral}
\affiliation{%
 \institution{Indian Institute of Technology Bombay}
 \city{Mumbai}
 \state{Maharashtra}
 \country{India}}
\email{rdabral@cse.iitb.ac.in}

\author{Ganesh Ramakrishnan}
\affiliation{%
 \institution{Indian Institute of Technology Bombay}
 \city{Mumbai}
 \state{Maharashtra}
 \country{India}}
\email{ganesh@cse.iitb.ac.in}

\title{LIGHTEN: Learning Interactions with Graph and Hierarchical TEmporal Networks for HOI in videos}

\begin{abstract}
  Analyzing the interactions between humans and objects from a video includes identification of the relationships between humans and the objects present in the video. It can be thought of as a specialized version of Visual Relationship Detection, wherein one of the objects must be a human. While traditional methods formulate the problem as inference on a sequence of video segments, we present a hierarchical approach, \model, to learn visual features to effectively capture spatio-temporal cues at multiple granularities in a video. Unlike current approaches, \model\  avoids using ground truth data like depth maps or 3D human pose, thus increasing generalization across non-RGBD datasets as well. Furthermore, we achieve the same using only the visual features, instead of the commonly used hand-crafted spatial features. We achieve state-of-the-art results in human-object interaction detection (88.9\% and 92.6\%) and anticipation tasks of CAD-120 and competitive results on image based HOI detection in V-COCO dataset, setting a new benchmark for visual features based approaches. Code for LIGHTEN is available at \textcolor{blue}{\url{https://github.com/praneeth11009/LIGHTEN-Learning-Interactions-with-Graphs-and-Hierarchical-TEmporal-Networks-for-HOI}}
\end{abstract}


\begin{teaserfigure}
  \includegraphics[width=\textwidth]{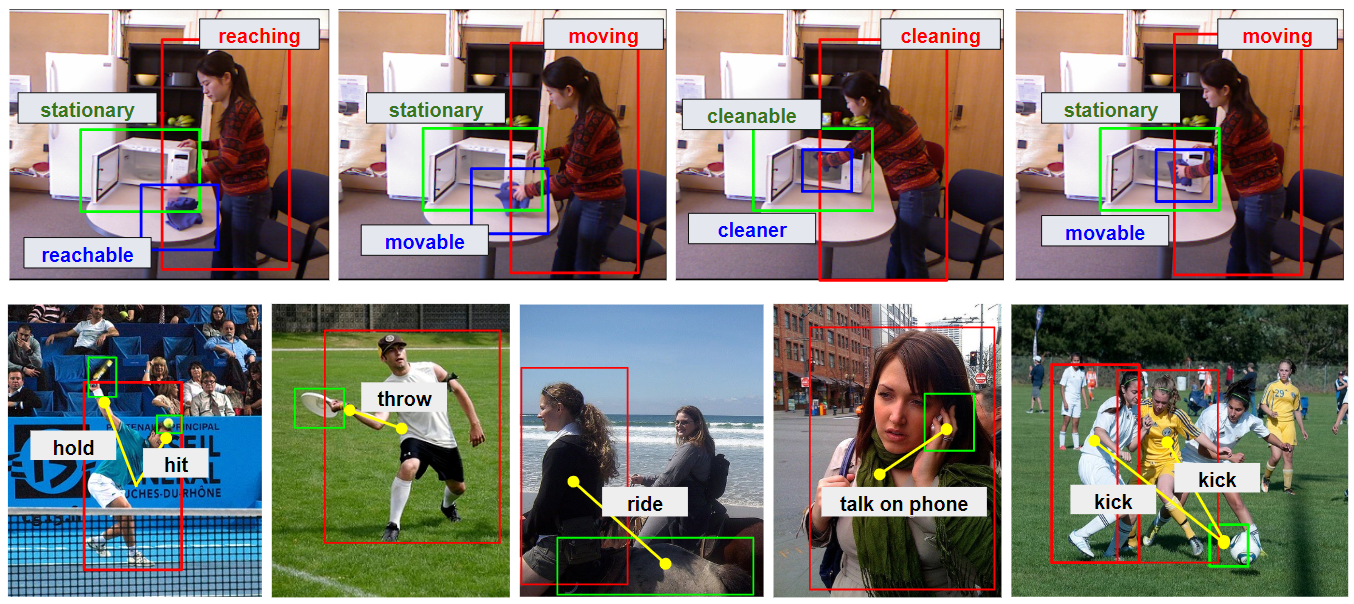} 
  \caption{Illustration of human-object interaction detection in video (CAD-120) and image (V-COCO) settings}
  \label{fig:teaser}
\end{teaserfigure}

\maketitle

\section{Introduction}
A key element of Scene Understanding is perception and interpretation of humans and the associated interactions. While human perception typically involves inferring the physical attributes about the humans (detection~\cite{dalal,Zhang_2018_ECCV, reploss, NIPS2015_5638}, poses~\cite{hourglass,Dabral:ECCV:2018,XNect_SIGGRAPH2020, DensePose, Varol_2017_CVPR,rishabh_multiperson}, shape~\cite{hmrKanazawa17,omran2018NBF, patel20tailornet, lazova3dv2019}, gaze~\cite{Xiong_2019_CVPR} etc.), interpreting humans involves reasoning about the finer details relating to human activity~\cite{action_in_still_images, actionbases, action_Liu_2020_CVPR, Nagrani_2020_CVPR, action_Zhang_2020_CVPR}, behaviour~\cite{EVA, Mittal_2020_CVPR}, human-object visual relationship detection~\cite{video_vrd, vrd_st_graph, vrd_feature_fusion, vrd_with_st_global_context, gated_st_energy_graph, annotating_objects_relations}, and human-object interactions~\cite{GPNN,PMFNet,video_vrd,vrd_st_graph,vrd_feature_fusion, vrd_with_st_global_context,gated_st_energy_graph,annotating_objects_relations}. In this work, we investigate the problem of identifying Human-Object Interactions in videos. Given a video stream, the goal is to identify the objects interacting with the humans while also estimating the kind of interaction, {\em eg.}, holding the cup, placing the bowl, moving the furniture, {\em etc.} The availability of such information can be crucial in understanding the finer details of human behaviour than in, say, action recognition. Such information has the potential to facilitate downstream applications like unmanned supermarkets, surgery documentation, robotics, {\em etc.}

While investigating into the HOI problem, we especially focus on video settings. There has been a significant amount of research on HOI with images~\cite{knowledgeGraph,GPNN,interactiveness,PMFNet}, thanks to the availability of V-COCO~\cite{vcoco} and HICO~\cite{hico} datasets. However, learning human-object interactions within videos is challenging and relatively less explored owing to multiple reasons. \textit{Firstly}, it requires the model to account for the changing orientations of objects in the scene with respect to the humans. This makes it difficult to extend the image-based approaches that use the RoI features of the union of human and object to the video setting. 
\textit{Secondly}, the unavailability of large scale video datasets (except CAD-120~\cite{cad120}) makes it difficult to train an HOI model that is generic, and performs well for in-the-wild videos. \textit{Finally}, the interaction definitions tend to become confusing when defined for a video, {\em e.g.}, \textit{placing} vs. \textit{moving} vs. \textit{reaching}, \textit{opening} a jar vs. \textit{closing} a jar, etc. In spite of these challenges, videos allow for exploiting temporal visual cues that are, otherwise, absent in images. 

Most existing methods are designed to work in either the image setting\cite{interactiveness,PMFNet}, or the video setting~\cite{atrcf,srnn} but not both. Recently, Qi et al.~\cite{GPNN} proposed a graph-parsing based method that fits into to both the settings. While the method indeed achieves state-of-the art results in video setting, it does so by using carefully designed and pre-computed hand-crafted features such as SIFT~\cite{sift} transforms, object centroids, 3D poses, object depths, etc. which were originally proposed in~\cite{cad120}. It is worth noting that these features were derived from the ground-truth data provided by the CAD-120 dataset. It is straightforward to see that using ground-truth based features for estimating HOI would not allow the method to perform well on in-the-wild videos because such features may either not be available (3D pose) or may be noisy and inconsistent across frames (object bounding boxes, centroids, {\em etc.})

With these caveats in mind, we propose a hybrid approach that uses Graph Convolutional Network (GCN) and hierarchical RNNs, \model, for detecting human-object interactions from videos that \textit{does not} rely on hand-crafted features. We use pure visual features derived from a re-trainable off-the-shelf network to represent the inputs to \model\ and demonstrate strong performance on the CAD-120 dataset. Furthermore, The proposed network is designed to leverage the spatio-temporal cues that are crucial to disambiguate confusing interactions. Specifically, we design a two-level architecture which, i) performs graph-based spatial embedding extraction from the video and learns temporal reasoning functions at the frame level, followed by ii) a segment level temporal network which learns inter-segment temporal cues from previous segments, for regressing the human sub-activities and object affordances. The temporal functions are designed to learn the temporal relationships between human-object pairs across the video.

Despite not using the ground truth based pre-computed features and in spite of the small amount of data available for training from videos, our visual input based model achieves state-of-the-art performance on sub-activity, affordance detection tasks, setting a strong baseline for the future of such methods. When used with the segment level pre-computed features, the segment-level temporal model of our proposal performs at par with the state-of-the-art methods. Finally, despite being designed for video-based tasks, our method also demonstrates competitive performance on the V-COCO dataset that corresponds to the image setting.

In summary, we make three contributions in this paper in the form of our model, \model : \textit{First}, we propose a generalizable, multi-level method for identifying Human-Object Interactions from videos. To the best of our knowledge, ours is the first that performs video-based HOI estimation purely from learnt visual features. \textit{Second}, we setup a new baseline for such methods as ours on CAD-120 dataset while also approaching competitive results with methods that are either purely image-based or purely video-based. \textit{Third}, we show how \model\ naturally lends itself to static, image-based settings. 

\section{Related Work}
Human-Object Interaction detection has been a well researched problem. In this section, we discuss the existing literature from two broad viewpoints: static (images) and dynamic (videos).\\
\indent \textbf{HOI from images}: Prior to deep learning, initial works on HOI from images were based on using hand-crafted features such as SIFT, HOG, etc. Among such works, Yao {\em et. al.}~\cite{actionbases} learned the bases of actions and parts to reason about HOI. Likewise, Hu {\em et. al.}~\cite{hu_exemplars} used HOI exemplars to model the spatial relationships between the human and the objects. A problem like Human-Object Interaction should be amenable to the use of structure based reasoning, by virtue of the fact that HOI requires detection of humans and objects and their spatial interactions that are expected to persist temporally.
Toward this, Yao {\em et. al.}~\cite{grouplet} define \textit{grouplets} as a feature encoder for capturing structural information, Delaitre {\em et. al.}~\cite{action_in_still_images} construct structure-aware feature representations that are trainable with an SVM. \\
\indent Recently, deep learning based methods, bolstered by the availability of large amounts of in-the-wild training data~\cite{vcoco,hico} have lead to significantly improved performance in HOI detection.  Among such methods, Li {\em et. al.}~\cite{interactiveness} proposed to learn the knowledge about the \textit{interactiveness} between the humans and object categories from HOI datasets and use this knowledge as a prior while performing HOI detection. For understanding the interactions, it has also been argued that human pose provides useful cues about the type of interaction. For example, a human \textit{opening} a jar will have a significantly different pose than when the human is \textit{reaching} for a jar. Several methods have attempted to leverage the human pose information in their pipelines. Wan {\em et. al.}~\cite{PMFNet} propose a pose-aware network architecture that employs a multi-level feature strategy, thereby dealing with the problem at three levels of granularity: overall interactions (covering both human and object), independent visual cues from the object and the human RoIs, and the fine-grained body part level features. Likewise, Xu {\em et. al.}~\cite{intend} use the human pose features in conjunction with the gaze estimates to discover human intentions, which are then used for HOI detection. Since the HOI problem is well-suited for graph-based representations, Graph Convolutional Networks have been regularly used to model the interactions. In this line of work, Xu {\em et. al.}~\cite{knowledgeGraph} propose to deal with long-tail HOI categories by modeling underlying regularities among verbs and objects. They do so by constructing a knowledge graph and enforcing similarity of graph embeddings derived from a GCN with visual feature embeddings derived from a CNN using a triplet-loss. Qi {\em et. al.}~\cite{GPNN} propose GPNN, a method that uses an iterative message passing framework on a parse graph comprising of verbs and objects as nodes. Our work is inspired by graph based methods in that we represent humans and objects as graph nodes and learn their interactions based on the image-based node features.\\

\indent \textbf{HOI from Video}: The HOI labels predicted in this task are typically indicative of an activity spanning over a period of time. Therefore, utilizing temporal cues in a video setting is naturally expected to provide important insights on the interactions and thereby benefit the HOI detection. Albeit less, there have also been significant contributions towards research on HOI detection in videos, mostly on the CAD-120 dataset. Koppula {\em et. al.}~\cite{cad120} proposed the dataset and introduced an MRF base formulation for handling spatio-temporal requirements. The authors hand-crafted a set of features for humans (pose, displacement of joints, {\em etc.}) and objects (3D centroids, transforms of SIFT matches between adjacent frames, {\em etc}). Instead of being used at the frame-level, these features, put together, represented a video segment as a whole. Since then, most existing methods (deep learning and traditional methods alike) work on the same segment level features. Qi {\em et. al.}~\cite{GPNN} extend their GPNN method for videos and construct a parse graph for every video segment using the segment level features to initialize the node and edge features in their parse graph. Likewise, Jain {\em et. al.}~\cite{srnn} design a spatio-temporal graph for performing structured predictions on a video consisting of multiple segments. Kopulla {\em et. al.}~\cite{atrcf} present ATCRF - a CRF based approach that models anticipatory trajectories of objects and humans. \\
\indent While there have been remarkable improvements over the years, we submit that there are two major areas for improvement. Firstly, avoiding over-dependence on such hand-crafted features, since the above approaches limit the scope for in-the-wild HOI detections. Such over-dependence has been averted in both textual~\cite{aaai2020} and image~\cite{aaai2018} domains and we take inspiration from such works. More often than not, the 3D poses or 3D centroids of objects (used as features) are either not available or are too erroneously estimated to be simply plugged into a model trained on hand-crafted features. Secondly, all the existing methods model temporal relations only between multiple \textit{segments} of a video. This may be, partly, because the hand-crafted features discussed above are defined for a segment as a whole. We believe that there is scope for exploring temporal cues even at a more fine-grained level, {\em viz.}, frame-level. Using image-based features facilitates the same. \\
We, therefore, propose an approach to model HOI relevant spatial-structures from every frame of a segment and further design a temporal aggregation regime using these frame level structures. Again, such aggregation strategies have provided to be extremely effective in other domains such a entity-linking~\cite{kdd2009,cods2016}. Deep-learning based computer vision models have enough representation power to be able to extract meaningful visual features from images or videos. Thus, our primary intent is to construct a model which can effectively learn hierarchical HOI embeddings at a fine-grained frame level as well as at a coarser segment level, using only visual attributes, and set a new baseline for human-object interaction detection in videos.

\begin{figure*}
  \includegraphics[width=\textwidth]{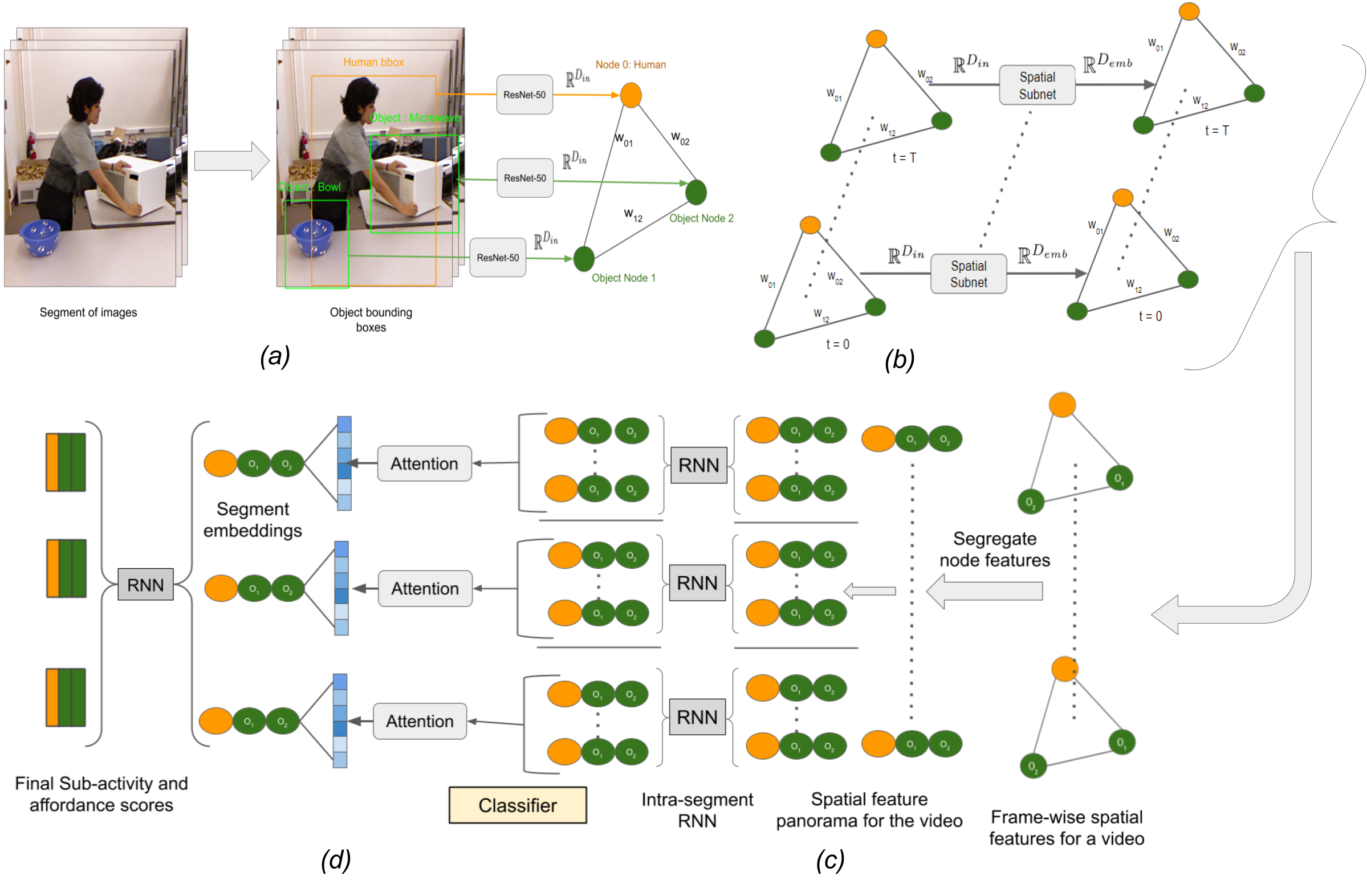}
  \caption{Overall pipeline in \model. Given an input video segment with T frames and bounding box coordinates of the humans and objects in every frame, we (a) first extract the visual features from ResNet-50. (b) These features are then processed in a per-frame fashion by a Spatial Subnet. (c) The graph structure is disentangled and temporal cues between frames in a segment are learnt from spatial features. (d) The frame-wise features are summarised into segment embeddings using attention mechanism and refined using inter-segment relations, to regress the human subactivities and object affordances.}
  \label{fig:pipeline}
\end{figure*}

 \section{Our Approach: \model}
In this section, we present our method, \model\ (Learning Interactions using Graphs and Hierarchical TEmporal Networks) for HOI detection on video. The HOI information in the videos can be  dealt with at 
two levels of granularity. The first, and the coarser, granularity corresponds to viewing the video as a sequence of segments, with each segment representing an atomic interaction. For example, a video may include a sequence of segments such as: \textit{reaching} for a jar, \textit{opening} the jar, and \textit{placing} the jar back. The second, and finer, granularity corresponds to dissecting each segment into its constituent frames. Lastly, the visual features at frame level provide crucial spatial cues about the possible interactions.

In \model, we attempt to exploit these well defined constructs and put them under consideration when choosing the architecture. The overall pipeline of \model\ is illustrated in Figure \ref{fig:pipeline}. 
  
 \subsection{The Proposed Learning Architecture}
 Given an input video $\mathcal{I} = \{I_1, I_2, \dots, I_T\}$ consisting of $T$ frames such that the video includes a single human and $N$ objects, our task is to regress human subactivities (placing, opening, {\em  etc.}), $H = \{H_0, H_1, \dots, H_M\}$ for the human and object affordances (placable, openable \textit{etc.}), $O = \{O_{0,0}, O_{0,1}, \dots, O_{N,M}\}$ for each of the $N$ objects and $M$ segments in the video. To this end, we propose a pipeline consisting of three stages: (i) the spatial subnet, (ii) the frame-level temporal subnet, and (iii) the segment-level temporal subnet.\\

The spatial subnet feeds on an input frame $I_t$ and learns a set of embeddings $\phi_t \in \mathbb{R}^{D_{emb}}$ for each human and $\theta_{n,t} \in \mathbb{R}^{D_{emb}}$ for each object. These per-frame, spatial embeddings are then fed to the \textit{frame-level} temporal subnet that churns out the corresponding spatio-temporal embeddings, $\Phi_t \in \mathbb{R}^{D_{emb}}$ and $\Theta_{n,t} \in \mathbb{R}^{D_{emb}}$, while also providing initial estimates of $H_m$ and $O_{n,m}$, where $m$ corresponds to the segment index, and $n$ corresponds to the object index. The frame-level spatio-temporal embeddings are then consolidated for each segment using an attention mechanism to produce $A^{\Phi}_m$ and $A^{\Theta}_{n,m}$, and passed on to \textit{segment-level} temporal subnet that produces the final outputs for the subactivity and affordance estimates.

To the best of our knowledge, \model\ is the first approach to detection of human-object interactions from videos that is completely pivoted on end-to-end learning. On the  contrary, majority of prior work~\cite{GPNN,srnn,atrcf} has dealt with the problem only at the segment level. Furthermore, previous work has derived spatial features not from the raw images, but from the ground-truth data like depth of the objects, pose of the human and objects, {\em etc.} It is easy to see that such a construction prohibits its use on any video for which depth information is unavailable.  Next, we now elaborate on each step of the pipeline.

\subsection{Spatial Subnet}
As just discussed, the sole job of the spatial subnet is to learn features relevant to the spatial ordering of the objects and the human. We model this task in a Graph Convolutional Network (GCN) setting which lends itself naturally to the task at hand. We define the graph $\mathcal{G} = (\mathcal{V}, \mathcal{E})$, where the nodes $\mathcal{V} = \{1, 2, \dots, N+1\}$ correspond to $N$ objects and one human and $\mathcal{E} = (p,q) \in \mathcal{V}\times\mathcal{V}$. \\
\indent We extract the node features $x_{v,t} \in \mathbb{R}^{D_{in}}$ corresponding to the $v^{th}$ node (human/object) of the $t^{th}$ frame by feeding the corresponding image crop $I_{v,t}$ to an off-the-shelf feature extractor $F$. Formally, $x_{v,t} = F(I_{v,t})$.  The edge weights are initialized to be 1 for human-object edges and 0 for the rest. The adjacency matrix is dynamically learnt while training the Spatial Subnet. \\

 A major challenge in GCN based formulation is to account for variability in the number of nodes across segments in a video. For example, a video may include the following segments: picking a bowl (1 object), moving the bowl (1 object), putting the bowl in the microwave (2 objects). Typically, this number varies from two nodes to six nodes.\\

\begin{figure}[h]
  \centering
\includegraphics[width=\linewidth]{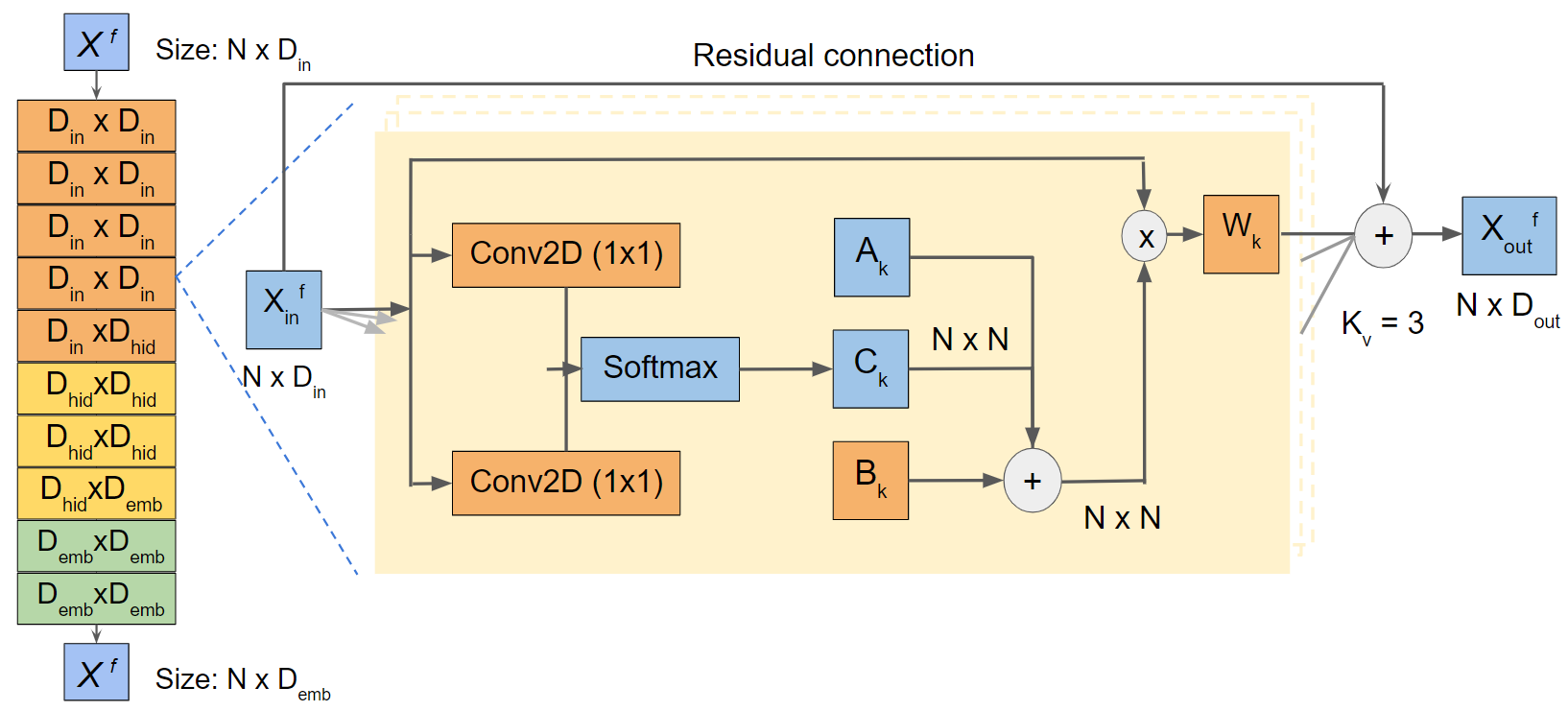}
\caption{Architecture of Spatial Subnet. Each block augments the adjacency matrix by a learnable correction, $B$, and a data-dependent course-correction, $C$. A residual connection is added to facilitate faster training of the model}\label{figure:AGCN}
\Description{}
\end{figure}

A trivial solution would be to design the GCN with a maximum number of nodes (six, in this case), initialize the unused nodes with zeros, and expect the network to learn to recognize the dummy nodes. This, however, leads to inferior results. To alleviate this issue, the network is designed to inherently learn course-corrections to the adjacency matrix. As depicted in Figure ~\ref{figure:AGCN}, every graph-convolution layer is followed by an update of the adjacency matrix which involves addition of the following two refinement components to the base adjacency matrix \textit{A}. The first component is a learnable additive matrix, \textit{B} that is learnt during the training process. The second component is a data-driven additive matrix, \textit{C} that is estimated uniquely for every input. This formulation has been inspired by the Adaptive Graph Convolution Network proposed in~\cite{2sagcn2019cvpr}. However, unlike~\cite{2sagcn2019cvpr}, we do not operate in the time dimension at the level of the GCN. \\
 
Formally, the Spatial Subnet, $S$ transforms the features corresponding to the $t^{th}$ frame as $\phi_t = S(x_{v,t})$ if $v$ is a human node and $\theta_t = S(x_{v,t})$ if $v$ corresponds to an object node. At the end of the Spatial Subnet, the network produces an intermediate feature set in $\mathbb{R}^{T\times (N+1)\times D_{emb}}$ space.

\subsection{Frame-level Temporal Subnet}
 Once the per-frame spatial features for the graph are extracted, (in the case of video data such as CAD-120) we process the graph features in time dimension, thus providing a feature-panorama of the entire segment. As discussed earlier, temporal reasoning occurs in two granularities - frame level and segment level. It is at this stage that we dis-integrate the graph structure of the network and construct individual feature sets for each node, aggregated over time. These frame-level embeddings are subjected to a bidirectional Recurrent Neural Network (RNN) which produces two outputs for every frame:\\
  
\indent For human nodes, given the input embeddings $\phi_t \in \mathbb{R}^{T\times N\times D_{emb}}$, the frame-level bidirectional-RNN outputs the estimates of human subactivity, $H_{m,t}$, and updates the recurrent embedding, $\Phi_t \in \mathbb{R}^{D_{emb}}$ for frame $t$ in segment $m$.   
Note, that while the learnt embeddings are further fed into the segment-level subnet, we also use them to classify subactivities and affordances for each frame to facilitate stronger supervision.
\indent For object nodes, we concatenate human node features along with the object node features and feed it to the frame-level RNN which outputs the estimates of object affordances $O_{n,m,t}$ and updates the corresponding recurrent embeddings, $\Theta_{n,t} \in \mathbb{R}^{D_{emb}}$\\

The aggregated activity and affordance classification scores at frame level are computed by taking a summation of the sequential frame-wise scores output by the RNN. Formally, the frame-level subactivity prediction can be written as: ${H}_{m} = softmax(\sum_t H_{m, t})$ 

\indent One key driver behind this form of segregated temporal aggregation for each object, as opposed to joint inference across all objects and humans is the variability in the number of objects in the scene. As such, we leave the job of inter-object relationship discovery to the spatial subnet and only exploit human-object correlations while making the temporal predictions.

\indent \textbf{Loss Functions:} We subject both the classifiers to standard Cross-Entropy losses  $\mathcal{L}_h$  and $\mathcal{L}_o$,
The overall loss is a weighted sum of the two losses and can be written as: 
\begin{equation*}
   \mathcal{L}  =  \mathcal{L}_h + \lambda \mathcal{L}_o
\end{equation*}

\subsection{Segment-level Temporal Subnet}
The previous subnet learns intra-segment temporal relations, but does not utilize the temporal information from the previous segments of the video, thus lacking wider context. With the segment-level subnet, we aim to learn inter-segment temporal cues by leveraging the context from previous segments of the video. We use another RNN to model these relations.\\

\indent \textbf{Attention Mechanism:} The input to the segment-level RNN is a sequence of embeddings, $A^{\Phi}_m$, corresponding to each segment for human nodes. We extract $A^{\Phi}_m$ by subjecting the frame-level embeddings, $\Phi_{m,t}$ to an attention network that produces a single embedding for a segment. Formally, $A^{\Phi}_m = \sum_t a_{t} * \Phi_{m,t}$, where $a_t$ are the attention weights produced by a Multi-Layered Perceptron (MLP). Similar construction follows for the derivation of $A^{\Theta}_m$.

An alternative to this approach could have been to use the embedding corresponding to the last time step, $\Phi_{m,T}$ as the input to the segment-level RNN. While it works well, we observed superior performance with attention-guided mechanism.

\indent The summarized sequence of segment embeddings is finally processed by an RNN, to leverage temporal dependencies from the previous segments for predicting human subactivity and object affordances for the current segment.

We use the same loss functions for classifiers at both frame-level and segment-level.\\

\begin{table}
\caption{{A comparison of our approach with the existing methods. Note that unlike ours, all the methods that we compare with have been trained on hand-crafted features derived from the ground-truth spatial attributes including 3D human pose, object centroids. We obtain the state-of-the-art results in both subactivity, affordance detection tasks while learning the embeddings from RGB data. Seg-RNN corresponds to segment-level RNN}}
\label{table:results}
\begin{tabular}{|p{0.25\textwidth}||p{0.06\textwidth}|p{0.1\textwidth}|}
\hline
& \multicolumn{2}{c||}{F1 Score in \%} \\\cline{2-3}
Method & Sub-activity &  Object Affordance\\
\hline
\hline
ATCRF~\cite{atrcf} & 80.4 & 81.5\\
S-RNN~\cite{srnn} & 83.2 & 88.7\\
S-RNN (multi-task)~\cite{srnn} & 82.4 & 91.1\\
GPNN~\cite{GPNN} & 88.9 & 88.8\\
\hline
\textbf{\model\ w/o Seg-RNN} & \textbf{85.9} & \textbf{88.9}\\
\textbf{\model\ (full model)} & \textbf{88.9} & \textbf{92.6}\\
\hline
\end{tabular}
\end{table}

\subsection{Implementation Details}
We now discuss implementation details from two vantage points: model and training.\\
\indent \textbf{Model:} Since the number of frames in a video segment may vary significantly, we uniformly sample a fixed number of frames, T, from the segment (for our experiments on CAD-120 dataset, we use T=20). We extract the RoI crops from each frame and reshape them to a  fixed size of $224\times224\times3$ (input dimension for ResNet). For our experiments, we explore the usage of ResNet-34, ResNet-50~\cite{resnet} as the feature extractors that produces 512-dimensional (2048 for ResNet-50) features for every node of the graph. Since we have limited data, we use the pre-trained ResNet features. In order to incorporate the information on positioning of humans and objects, we append normalized bounding box coordinates of human/objects to their respective visual node features. We use a hidden, output feature dimensions of 512 for the graph convolutional network of Spatial Subnet. \\
\indent \textbf{Training:}
We use the PyTorch deep learning framework for implementing \model. During training, we set $\lambda=2$ for the overall loss. We use the Adam~\cite{adam} optimizer with initial learning rate of $2\times10^{-5}$, learning rate decay factor of 0.8, and decay step size of 10 epochs. We train \model\ for a total of 300 epochs on Nvidia RTX 2080Ti GPU. We performed a hyper-parameter sweep to empirically obtain these configurations. The entire model is trained in two steps. Firstly, the model up to frame-level temporal subnet is trained by aggregating classification scores from the $T$ frames of the segment. Finally, the entire model including the segment-level subnet, is trained in an end-to-end fashion, after initializing the parameters from the pre-trained frame-level model.

 \section{Experiments}
\subsection{Datasets}
We evaluate \model\ for the task of Human-Object Interaction detection on two datasets, {\em viz.}, i) CAD-120~\cite{cad120} ,and ii) V-COCO~\cite{vcoco}. \\

\indent \textbf{CAD-120:} The CAD-120 dataset is a video dataset with 120 RGB-D videos of 4 subjects performing 10 daily indoor activities (\textit{e.g., making cereal, microwaving food}). Each activity is a sequence of video segments involving finer-level activities. In each video segment, the human is annotated with an activity label from a set of 10 sub-activity classes \textit{(e.g., reaching, pouring)} and each object is annotated with an affordance label from a set of 12 affordance classes (\textit{e.g., pourable, movable}). The frame-length of each segment ranges from 22 to a little over 150 frames. \\

\indent The metrics used for evaluating \model\ on the human-object interaction tasks of CAD-120 dataset are: i) sub-activity F1-score, and ii) object affordance F1-score computed for human sub-activity classification and object affordance classification. The dataset, in addition to providing the images and HOI annotations, additionally provides depth maps, 3D pose information and segment-level hand-crafted spatial features. We do not make use of any additional data except the 2D bounding box of the objects and humans, and aim to learn the segment embeddings from RGB data only.\\

\indent \textbf{V-COCO:} Crafted as a subset of the MS-COCO~\cite{coco_dataset} dataset,  V-COCO is an image dataset that provides annotations of Action labels for edges between human and object. There are 26 action classes.

\subsection{Quantitative Evaluation}
\subsubsection{Evaluation on the CAD-120 dataset: }
 The performance of \model\ is evaluated in two experimental setups. i) In the first setup, we pick the labels predicted directly from the output sequence at the frame-level subnet. In the second setup, ii) the subactivity and affordances are predicted after incorporating the segment-level RNN. In each of these two experiments, we train \model\  separately for the tasks of HOI detection and HOI anticipation. In all the experiments, the video data we provide as input to \model\ is: i) RGB frames of the video ii) bounding boxes of human and object in the frames of video.
 
\begin{figure}[h]
  \centering
\includegraphics[width=\linewidth]{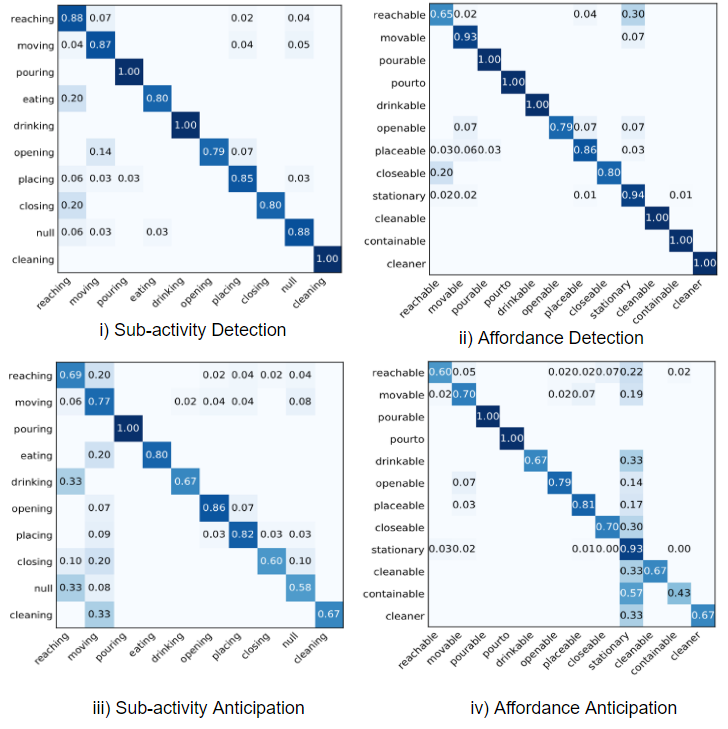}
\caption{Confusion matrices for human-object interaction detection setting -- (i), (ii) -- and anticipation setting -- (iii), (iv) -- on CAD120 dataset. It is worth  noting that most of the confusion occurs in visually similar categories, e.g.  closing vs. reaching and opening vs. moving}\label{figure:conf}
\Description{} 
\end{figure}

 We tabulate the results of our approach in Table \ref{table:results}. As the numbers suggest,  we achieve state-of-the-art performance with sub-activity detection F1 score of \textbf{88.9} and affordance detection F1 score of \textbf{92.6}. we also achieve an F1 score of \textbf{76.4} in human subactivity anticipation task, outperforming previous methods, and an F1 score of \textbf{78.8} in affordance anticipation task. To the best of our knowledge, all previous works on the task of human-object interaction in CAD-120, use the hand-crafted features provided by CAD-120 dataset. So we believe that this experiment is the only one which bypasses the usage of the handcrafted features and relies only on 2D video data, while achieving improved performance. \\
 We compare our method against the existing works on CAD-120: ATCRF~\cite{atrcf}, S-RNN~\cite{srnn}, and GPNN~\cite{GPNN}.\\
 \indent \textbf{Confusion Matrix}: The confusion matrices for both detection and anticipation tasks are displayed at Figure~\ref{figure:conf}. Every row of a confusion matrix indicates the prediction distribution of various node samples of that ground truth class. From the confusion matrix for affordance detection, it is evident that most of the false predictions of object nodes are due to misinterpretation of object as stationary. This is especially prevalent in the affordance class \textit{reachable}, because the human is usually far from the object during the sub-activity \textit{reaching}.

\begin{table}[h]
\caption{{A comparison of LIGHTEN on image-based HOI detection on V-COCO dataset }}
\label{table:vcoco_results}
\begin{tabular}{|p{0.3\textwidth}|p{0.13\textwidth}|}
\hline
Method & Role mAP score\\
\hline
\hline
Gupta et al. ~\cite{vcoco} & 31.8\\
InteractNet ~\cite{interact_net} & 40.0\\
GPNN ~\cite{GPNN} & 44.0\\
Li et al. ~\cite{interactiveness} & 48.6\\
PMFNet~\cite{PMFNet} & 52.0\\
\hline
LIGHTEN for image HOI & 38.28 \\
\hline
\end{tabular}
\end{table}

\subsubsection{Evaluation on V-COCO dataset}
Although our method is designed to leverage temporal cues within a video setting, we test our method on V-COCO dataset by setting T = 1. We observe the role mAP score of 38.28 which, although not close to the state-of-the-art method, achieves well reasonable performance without bells and whistles. We believe that an explanation for the sub-parity of our results is that in the absence of temporal cues, the spatial GCN is significantly shallower than other works and leads to inferior results. We provide a detailed comparison with other methods in Table ~\ref{table:vcoco_results}.

\subsection{Qualitative Evaluation}
\begin{figure}[h]
  \centering
\includegraphics[width=\linewidth]{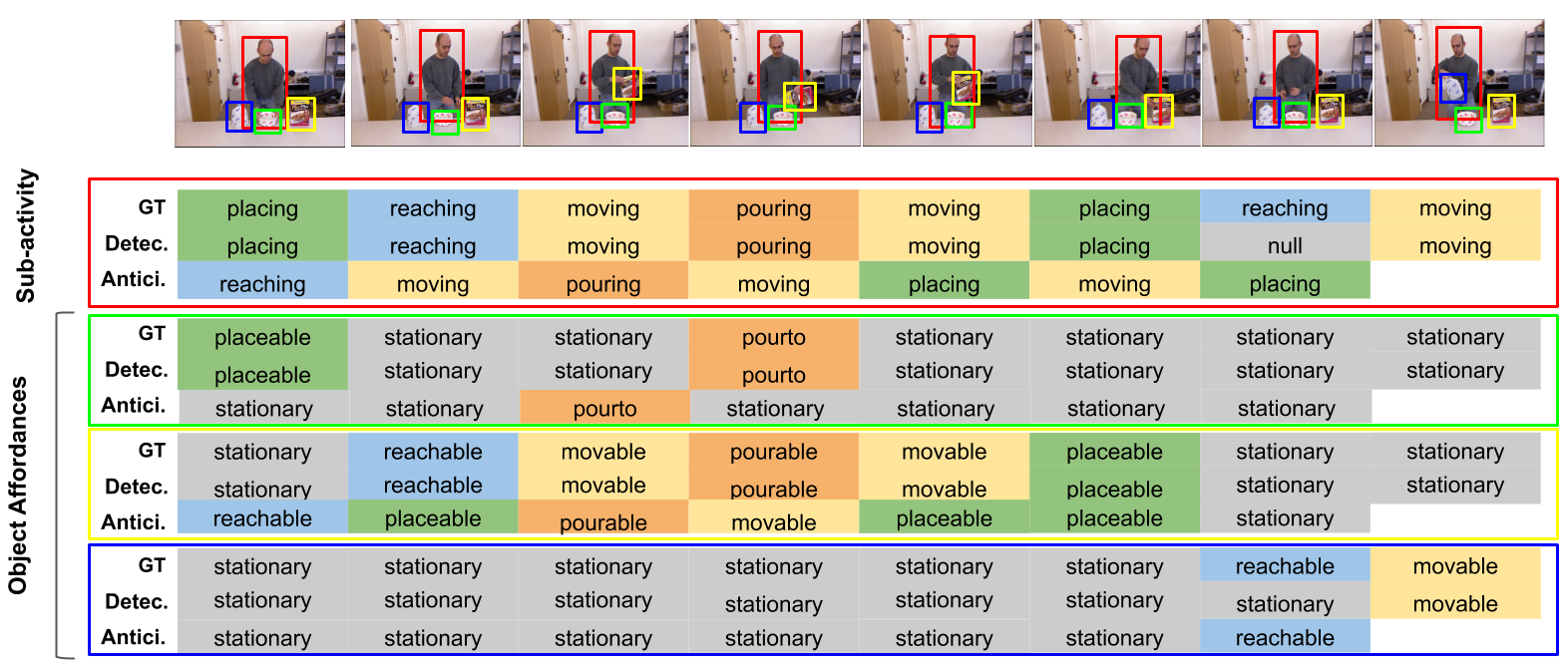}
\caption{Human Object Interaction Detection and Anticipation results on a video of activity "Making cereal" from the CAD-120 dataset.The nodes here are the human and three objects: i) bowl ii) milk iii) box. The object affordance predictions in the figure are for the objects in this order from top to bottom. Predictions are highlighted a border of same color (red for human, green for bowl, blue for milk, and yellow for box) as the human/object's bounding box in image. We show predictions for 8 segments of the video. The anticipation labels shown along with each segment are the labels anticipated for the upcoming segment.}
\label{figure:demo}
\Description{}
\end{figure}

\begin{figure}[h]
  \centering
\includegraphics[width=\linewidth]{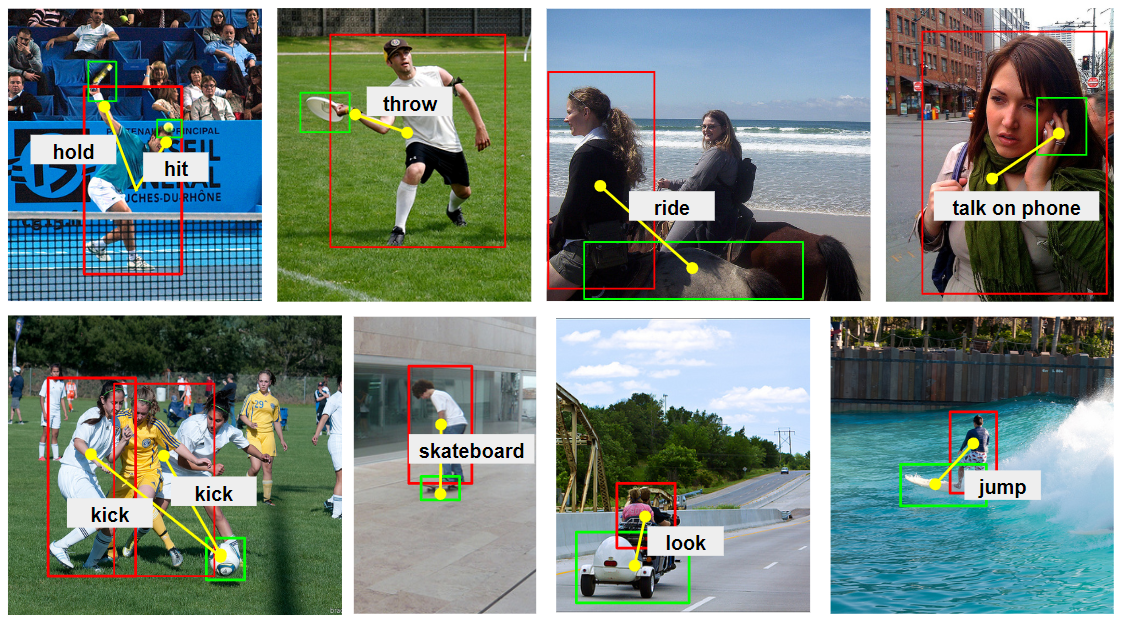}
\caption{ Detections of human-object action labels in test images of VCOCO. We report our failure cases on the last two images (bottom right). The rest are correct predictions.}
\label{figure:vcoco_test}
\Description{}
\end{figure}

We provide some qualitative evaluation of \model\ on CAD-120 dataset in Figure~\ref{figure:demo}. We see that while the HOI detections have been achieved accurately, there remains ambiguity among some classes during the anticipation task. \\
\indent Figure~\ref{figure:vcoco_test} demonstrates some positive and negative cases of detection of edge action labels of human-object pairs for test images on V-COCO. In the absence of temporal context, the method resorts to associating visual cues to spatial cues, thus not being able to disambiguate whether a person is \textit{sitting} on a car or \textit{looking} at the same car.

\begin{table}
\caption{A comparison of \model\ on anticipation task. Our approach achieves state-of-the-art results on human subactivity anticipation whereas performs competitively on object affordance anticipation.}
\label{table:anticipation}
\begin{tabular}{|p{0.29\textwidth}|p{0.06\textwidth}|p{0.08\textwidth}|}
\hline
& \multicolumn{2}{c||}{F1 Score in \%} \\\cline{2-3}
Method & \shortstack{Sub-\\activity} & \shortstack{Object\\Affordance} \\
\hline
\hline
ATCRF~\cite{atrcf} & 37.9 & 36.7\\
S-RNN~\cite{srnn} & 62.3 & 80.7\\
S-RNN (multi-task)~\cite{srnn} & 65.6 & 80.9\\
GPNN~\cite{GPNN} & 75.6 & 81.9\\
\hline
\textbf{\model\ w/o Segment-level subnet} & \textbf{73.2} & \textbf{77.6} \\
\textbf{\model\ (full model)} & \textbf{76.4} & \textbf{78.8} \\
\hline
\end{tabular}
\end{table}

\begin{table*}
    \caption{Ablation experiments of the impact of design choices on subactivity and object affordance detection. Seg-RNN refers to segment-level RNN and vanilla GCN refers to GCN without adjacency matrix refinement.}
    \begin{tabular}{|l||c|c|}
        \hline
        Experiment & Human Subactivity & Object Affordance \\
        \hline
        \hline
        \model\ w/o seg-RNN w/o spatial GCN & 61.5 & 78.6 \\
        \hline
        \model\ w/o seg-RNN with vanilla GCN block w/o residual connections & 70.3 & 61.3 \\
        \model\ w/o seg-RNN with vanilla GCN block with residual connections & 79.3 & 83.1 \\
        \hline
        \model\ w/o seg-RNN with MLP for frame-level temporal learning & 84.1 & 85.0 \\
        \model\ w/o seg-RNN w/o appending human node features to object nodes & 85.2 & 84.6 \\
        \model\ w/o seg-RNN  & 85.9 & 88.9 \\
        \hline
        \model\ w/o attention & 83.5 & 86.1 \\ 
        \model\ w/o seg-RNN with MLP for segment level temporal function  & 89.7 & 90.5 \\ 
        \hline
        Seg-RNN on hand-crafted features & 85.3 & 91.6 \\
        \model\  & 88.9 & 92.6 \\
        \hline
    \end{tabular}
    \label{tab:ablations}
\end{table*}

\subsection{Ablation Study}
We now discuss the contributions of various components to the final performance and their relevance to Human-Object Interaction detection. 
\subsubsection{Role of Graph Convolutions in Spatial Subnet: } Firstly, to verify the effectiveness of spatial graph convolution module, we designed an experiment where the image features from the backbone are directly passed to the frame-level model. We observed a significant degradation in performance in the absence of spatial GCN. While exploring variants of Graph Convolutional Networks, we also explored using a vanilla GCN network with basic graph convolution (GCN) layers as a baseline. As an extension to the basic GCN, we add a residual connection, similar to ~\cite{resnet}, which allows the input features to retain their initial behaviour. Using a residual connection brings an improvement in performance of GCN, as illustrated in Table~\ref{tab:ablations}. Further adding adaptive and data-dependent components to adjacency matrix, in a fashion similar to ~\cite{2sagcn2019cvpr}, also improves subactivity and affordance prediction, largely due to the ability to learn the inter-node edge weights.

\subsubsection{Role of human node features in affordance prediction: } In the temporal subnet, we concatenate human node features along with object node features for the frame and segment level RNNs. We observed significant improvement in performance on object affordance detection (88.6\% vs 84.6\%) due to human node features. This improvement can be attributed to the high correlation between the human sub-activity and affordances of active objects (objects which are not stationary). 

\subsubsection{Role of RNN in frame-level temporal subnet: } As a baseline for classification at frame-level subnet, we experimented with alternative temporal aggregation models. Specifically, we built an MLP network to obtain classification scores from spatial features concatenated across temporal dimension for each node separately. However, due to higher parameter count in MLP network, the model is prone to over-fitting, and therefore has a lower performance, which is evident from Table~\ref{tab:ablations}. 
\subsubsection{Role of segment-level temporal learning: } Even though subactivity and affordance labels are predicted for every single segment, there are significant inter-dependencies between the activity in a segment and activities in previous segments. As an illustrative example, in the following sequence of three segments in a video: \textit{reaching} for a jar, \textit{moving} the jar, and \textit{placing} the jar back, knowledge on the activities in first two segments can greatly improve the prediction of activity in the third segment. Using a temporal sequence processing network like an RNN after the frame-level aggregation step leverages these inter-segment dependencies and achieves a significant improvement in performance as compared to prediction at frame-level temporal subnet.
\subsubsection{Role of attention-mechanism in computing segment embedding: } We implemented two simpler baseline approaches to evaluate the use of attention weighting for frames. These approaches include i) using features corresponding to last frame in the output sequence of RNN ii) stitching the features across frames and regressing a segment embedding using MLP. Using the embedding corresponding to the last frame limits the representation power of the segment-level embedding $\Phi_m$. Using an MLP has the disadvantage of over-fitting and has an impact on object affordance detection as evident from the Table~\ref{tab:ablations}. 
\subsubsection{Evaluating the feature learning process:} To measure the effectiveness of the hierarchical learning mechanism, we design an experiment where we feed the hand-crafted, segment-level features to segment-level RNN, instead of the visual embeddings learnt by the attention mechanism. The learnt visual features achieve a better performance than the hand crafted features, particularly for the more difficult case of human subactivity detection (85.3\% vs 88.9\%), thereby justifying the effectiveness of the proposed method in capturing the spatio-temporal relations from RGB video data.

\section{Conclusion} In this paper, we proposed a two-step hierarchical approach for identifying Human-Object Interaction in videos. In the first step, we model the local interactions between humans and objects at a frame-level, while in the second step, we generate a segment-level embedding using the frame-level embeddings, and then refine them using the embeddings from previous segments. The embeddings are modelled through a graph structure, where the subject and object serve as nodes in a scene. Our approach  is easily extendable to other videos for the task of HOI, where depth information and 3D pose information is not available. Our approach sets a new benchmark for Human-Object Interaction detection in videos with visual information.

\section*{Acknowledgements \label{sec:ack}}
We are grateful to IBM Research, India (specifically the IBM AI Horizon Networks - IIT Bombay initiative) for their support and sponsorship. Rishabh Dabral has also been supported by Qualcomm Innovation Fellowship 2019.
\bibliographystyle{ACM-Reference-Format}
\balance
\bibliography{sample-base}


\begin{thebibliography}{50}


\ifx \showCODEN    \undefined \def \showCODEN     #1{\unskip}     \fi
\ifx \showDOI      \undefined \def \showDOI       #1{#1}\fi
\ifx \showISBNx    \undefined \def \showISBNx     #1{\unskip}     \fi
\ifx \showISBNxiii \undefined \def \showISBNxiii  #1{\unskip}     \fi
\ifx \showISSN     \undefined \def \showISSN      #1{\unskip}     \fi
\ifx \showLCCN     \undefined \def \showLCCN      #1{\unskip}     \fi
\ifx \shownote     \undefined \def \shownote      #1{#1}          \fi
\ifx \showarticletitle \undefined \def \showarticletitle #1{#1}   \fi
\ifx \showURL      \undefined \def \showURL       {\relax}        \fi
\providecommand\bibfield[2]{#2}
\providecommand\bibinfo[2]{#2}
\providecommand\natexlab[1]{#1}
\providecommand\showeprint[2][]{arXiv:#2}

\bibitem[\protect\citeauthoryear{Chao, Wang, He, Wang, and Deng}{Chao
  et~al\mbox{.}}{2015}]%
        {hico}
\bibfield{author}{\bibinfo{person}{Y.W. Chao}, \bibinfo{person}{Z. Wang},
  \bibinfo{person}{Y. He}, \bibinfo{person}{J. Wang}, {and} \bibinfo{person}{J.
  Deng}.} \bibinfo{year}{2015}\natexlab{}.
\newblock \showarticletitle{HICO: A benchmark for recognizing human-object
  interactions in images}. In \bibinfo{booktitle}{\emph{ICCV}}.
\newblock


\bibitem[\protect\citeauthoryear{Chatterjee, Ramakrishnan, and
  Sarawagi}{Chatterjee et~al\mbox{.}}{2020}]%
        {aaai2020}
\bibfield{author}{\bibinfo{person}{Oishik Chatterjee}, \bibinfo{person}{Ganesh
  Ramakrishnan}, {and} \bibinfo{person}{Sunita Sarawagi}.}
  \bibinfo{year}{2020}\natexlab{}.
\newblock \showarticletitle{Robust Data Programming with Precision-guided
  Labeling Functions}. In \bibinfo{booktitle}{\emph{AAAI}}.
\newblock


\bibitem[\protect\citeauthoryear{{Dabral}, {Gundavarapu}, {Mitra}, {Sharma},
  {Ramakrishnan}, and {Jain}}{{Dabral} et~al\mbox{.}}{2019}]%
        {rishabh_multiperson}
\bibfield{author}{\bibinfo{person}{R. {Dabral}}, \bibinfo{person}{N.~B.
  {Gundavarapu}}, \bibinfo{person}{R. {Mitra}}, \bibinfo{person}{A. {Sharma}},
  \bibinfo{person}{G. {Ramakrishnan}}, {and} \bibinfo{person}{A. {Jain}}.}
  \bibinfo{year}{2019}\natexlab{}.
\newblock \showarticletitle{Multi-Person 3D Human Pose Estimation from
  Monocular Images}. In \bibinfo{booktitle}{\emph{3DV}}.
\newblock


\bibitem[\protect\citeauthoryear{Dabral, Mundhada, Kusupati, Afaque, Sharma,
  and Jain}{Dabral et~al\mbox{.}}{2018}]%
        {Dabral:ECCV:2018}
\bibfield{author}{\bibinfo{person}{Rishabh Dabral}, \bibinfo{person}{Anurag
  Mundhada}, \bibinfo{person}{Uday Kusupati}, \bibinfo{person}{Safeer Afaque},
  \bibinfo{person}{Abhishek Sharma}, {and} \bibinfo{person}{Arjun Jain}.}
  \bibinfo{year}{2018}\natexlab{}.
\newblock \showarticletitle{Learning 3D Human Pose from Structure and Motion}.
  In \bibinfo{booktitle}{\emph{ECCV}}.
\newblock


\bibitem[\protect\citeauthoryear{Dalal and Triggs}{Dalal and Triggs}{2005}]%
        {dalal}
\bibfield{author}{\bibinfo{person}{Navneet Dalal} {and} \bibinfo{person}{Bill
  Triggs}.} \bibinfo{year}{2005}\natexlab{}.
\newblock \showarticletitle{Histograms of Oriented Gradients for Human
  Detection}. In \bibinfo{booktitle}{\emph{CVPR}}.
\newblock


\bibitem[\protect\citeauthoryear{Delaitre, Sivic, and Laptev}{Delaitre
  et~al\mbox{.}}{2011}]%
        {action_in_still_images}
\bibfield{author}{\bibinfo{person}{V. Delaitre}, \bibinfo{person}{J. Sivic},
  {and} \bibinfo{person}{I. Laptev}.} \bibinfo{year}{2011}\natexlab{}.
\newblock \showarticletitle{Learning person-object interactions for action
  recognition in still images}. In \bibinfo{booktitle}{\emph{NIPS}}.
\newblock


\bibitem[\protect\citeauthoryear{Gkioxari, Girshick, Dollar, and He}{Gkioxari
  et~al\mbox{.}}{2018}]%
        {interact_net}
\bibfield{author}{\bibinfo{person}{Georgia Gkioxari}, \bibinfo{person}{Ross
  Girshick}, \bibinfo{person}{Piotr Dollar}, {and} \bibinfo{person}{Kaiming
  He}.} \bibinfo{year}{2018}\natexlab{}.
\newblock \showarticletitle{Detecting and recognizing human-object
  interactions.}. In \bibinfo{booktitle}{\emph{CVPR}}.
\newblock


\bibitem[\protect\citeauthoryear{G\"uler, Neverova, and Kokkinos}{G\"uler
  et~al\mbox{.}}{2018}]%
        {DensePose}
\bibfield{author}{\bibinfo{person}{Riza~Alp G\"uler}, \bibinfo{person}{Natalia
  Neverova}, {and} \bibinfo{person}{Iasonas Kokkinos}.}
  \bibinfo{year}{2018}\natexlab{}.
\newblock \showarticletitle{DensePose: Dense Human Pose Estimation In The
  Wild}. In \bibinfo{booktitle}{\emph{CVPR}}.
\newblock


\bibitem[\protect\citeauthoryear{Gupta and Malik}{Gupta and Malik}{2015}]%
        {vcoco}
\bibfield{author}{\bibinfo{person}{Saurabh Gupta} {and}
  \bibinfo{person}{Jitendra Malik}.} \bibinfo{year}{2015}\natexlab{}.
\newblock \showarticletitle{Visual Semantic Role Labeling}.
\newblock In \bibinfo{booktitle}{\emph{arXiv preprint arXiv:1505.04474}}.
\newblock


\bibitem[\protect\citeauthoryear{He, Zhang, Ren, and Sun}{He
  et~al\mbox{.}}{2016}]%
        {resnet}
\bibfield{author}{\bibinfo{person}{Kaiming He}, \bibinfo{person}{Xiangyu
  Zhang}, \bibinfo{person}{Shaoqing Ren}, {and} \bibinfo{person}{Jian Sun}.}
  \bibinfo{year}{2016}\natexlab{}.
\newblock \showarticletitle{Deep Residual Learning for Image Recognition}. In
  \bibinfo{booktitle}{\emph{CVPR}}.
\newblock


\bibitem[\protect\citeauthoryear{Hu, Zheng, Lai, Gong, and Xiang}{Hu
  et~al\mbox{.}}{2013}]%
        {hu_exemplars}
\bibfield{author}{\bibinfo{person}{J.F. Hu}, \bibinfo{person}{W.S. Zheng},
  \bibinfo{person}{J. Lai}, \bibinfo{person}{S. Gong}, {and}
  \bibinfo{person}{T. Xiang}.} \bibinfo{year}{2013}\natexlab{}.
\newblock \showarticletitle{Recognising human-object interaction via exemplar
  based modelling}. In \bibinfo{booktitle}{\emph{ICCV}}.
\newblock


\bibitem[\protect\citeauthoryear{Jain, Zamir, Savarese, and Saxena}{Jain
  et~al\mbox{.}}{2016}]%
        {srnn}
\bibfield{author}{\bibinfo{person}{A. Jain}, \bibinfo{person}{A.R. Zamir},
  \bibinfo{person}{S. Savarese}, {and} \bibinfo{person}{A Saxena}.}
  \bibinfo{year}{2016}\natexlab{}.
\newblock \showarticletitle{Structural-RNN: Deep learning on spatio-temporal
  graphs}. In \bibinfo{booktitle}{\emph{CVPR}}.
\newblock


\bibitem[\protect\citeauthoryear{Kanazawa, Black, Jacobs, and Malik}{Kanazawa
  et~al\mbox{.}}{2018}]%
        {hmrKanazawa17}
\bibfield{author}{\bibinfo{person}{Angjoo Kanazawa},
  \bibinfo{person}{Michael~J. Black}, \bibinfo{person}{David~W. Jacobs}, {and}
  \bibinfo{person}{Jitendra Malik}.} \bibinfo{year}{2018}\natexlab{}.
\newblock \showarticletitle{End-to-end Recovery of Human Shape and Pose}. In
  \bibinfo{booktitle}{\emph{CVPR}}.
\newblock


\bibitem[\protect\citeauthoryear{Kingma and Ba}{Kingma and Ba}{2015}]%
        {adam}
\bibfield{author}{\bibinfo{person}{Diederik~P. Kingma} {and}
  \bibinfo{person}{Jimmy Ba}.} \bibinfo{year}{2015}\natexlab{}.
\newblock \showarticletitle{Adam: A Method for Stochastic Optimizations}. In
  \bibinfo{booktitle}{\emph{ICLR}}.
\newblock


\bibitem[\protect\citeauthoryear{Koppula, Gupta, and Saxena}{Koppula
  et~al\mbox{.}}{2013}]%
        {cad120}
\bibfield{author}{\bibinfo{person}{H.S. Koppula}, \bibinfo{person}{R. Gupta},
  {and} \bibinfo{person}{A. Saxena}.} \bibinfo{year}{2013}\natexlab{}.
\newblock \showarticletitle{Learning human activities and object affordances
  from RGB-D videos}. In \bibinfo{booktitle}{\emph{The International Journal of
  Robotics Research}}.
\newblock


\bibitem[\protect\citeauthoryear{Koppula and Saxena}{Koppula and
  Saxena}{2016}]%
        {atrcf}
\bibfield{author}{\bibinfo{person}{H.S. Koppula} {and} \bibinfo{person}{A.
  Saxena}.} \bibinfo{year}{2016}\natexlab{}.
\newblock \showarticletitle{Anticipating human activities using object
  aﬀordances for reactive robotic response}. In
  \bibinfo{booktitle}{\emph{TPAMI}}.
\newblock


\bibitem[\protect\citeauthoryear{Kulkarni, Agarwal, Shah, Rathod, and
  Ramakrishnan}{Kulkarni et~al\mbox{.}}{2016}]%
        {cods2016}
\bibfield{author}{\bibinfo{person}{Ashish Kulkarni}, \bibinfo{person}{Kanika
  Agarwal}, \bibinfo{person}{Pararth Shah}, \bibinfo{person}{Sunny~Raj Rathod},
  {and} \bibinfo{person}{Ganesh Ramakrishnan}.}
  \bibinfo{year}{2016}\natexlab{}.
\newblock \showarticletitle{Learning to Collectively Link Entities}. In
  \bibinfo{booktitle}{\emph{Proceedings of the 3rd {IKDD} Conference on Data
  Science, {CODS}}}.
\newblock


\bibitem[\protect\citeauthoryear{Kulkarni, Uppalapati, Singh, and
  Ramakrishnan}{Kulkarni et~al\mbox{.}}{2018}]%
        {aaai2018}
\bibfield{author}{\bibinfo{person}{Ashish Kulkarni},
  \bibinfo{person}{Narasimha~Raju Uppalapati}, \bibinfo{person}{Pankaj Singh},
  {and} \bibinfo{person}{Ganesh Ramakrishnan}.}
  \bibinfo{year}{2018}\natexlab{}.
\newblock \showarticletitle{An Interactive Multi-Label Consensus Labeling Model
  for Multiple Labeler Judgments}. In \bibinfo{booktitle}{\emph{AAAI}},
  \bibfield{editor}{\bibinfo{person}{Sheila~A. McIlraith} {and}
  \bibinfo{person}{Kilian~Q. Weinberger}} (Eds.). \bibinfo{publisher}{{AAAI}
  Press}.
\newblock


\bibitem[\protect\citeauthoryear{Kulkarni, Singh, Ramakrishnan, and
  Chakrabarti}{Kulkarni et~al\mbox{.}}{2009}]%
        {kdd2009}
\bibfield{author}{\bibinfo{person}{Sayali Kulkarni}, \bibinfo{person}{Amit
  Singh}, \bibinfo{person}{Ganesh Ramakrishnan}, {and} \bibinfo{person}{Soumen
  Chakrabarti}.} \bibinfo{year}{2009}\natexlab{}.
\newblock \showarticletitle{ACM SIGKDD},
  \bibfield{editor}{\bibinfo{person}{John F.~Elder IV},
  \bibinfo{person}{Fran{\c{c}}oise Fogelman{-}Souli{\'{e}}},
  \bibinfo{person}{Peter~A. Flach}, {and} \bibinfo{person}{Mohammed~Javeed
  Zaki}} (Eds.).
\newblock


\bibitem[\protect\citeauthoryear{Lazova, Insafutdinov, and Pons-Moll}{Lazova
  et~al\mbox{.}}{2019}]%
        {lazova3dv2019}
\bibfield{author}{\bibinfo{person}{Verica Lazova}, \bibinfo{person}{Eldar
  Insafutdinov}, {and} \bibinfo{person}{Gerard Pons-Moll}.}
  \bibinfo{year}{2019}\natexlab{}.
\newblock \showarticletitle{360-Degree Textures of People in Clothing from a
  Single Image}. In \bibinfo{booktitle}{\emph{3DV}}.
\newblock


\bibitem[\protect\citeauthoryear{Li, Zhou, Huang, Xu, Ma, Fang, Wang, and
  Lu}{Li et~al\mbox{.}}{2019}]%
        {interactiveness}
\bibfield{author}{\bibinfo{person}{Yong-Lu Li}, \bibinfo{person}{Siyuan Zhou},
  \bibinfo{person}{Xijie Huang}, \bibinfo{person}{Liang Xu},
  \bibinfo{person}{Ze Ma}, \bibinfo{person}{Hao-Shu Fang},
  \bibinfo{person}{Yan-Feng Wang}, {and} \bibinfo{person}{Cewu Lu}.}
  \bibinfo{year}{2019}\natexlab{}.
\newblock \showarticletitle{Transferable interactiveness prior for human-object
  interaction detection}. In \bibinfo{booktitle}{\emph{CVPR}}.
\newblock


\bibitem[\protect\citeauthoryear{Lin, Maire, Belongie, Hays, Perona, Ramanan,
  Doll{\'a}r, and Zitnick}{Lin et~al\mbox{.}}{2014}]%
        {coco_dataset}
\bibfield{author}{\bibinfo{person}{Tsung-Yi Lin}, \bibinfo{person}{Michael
  Maire}, \bibinfo{person}{Serge Belongie}, \bibinfo{person}{James Hays},
  \bibinfo{person}{Pietro Perona}, \bibinfo{person}{Deva Ramanan},
  \bibinfo{person}{Piotr Doll{\'a}r}, {and} \bibinfo{person}{C~Lawrence
  Zitnick}.} \bibinfo{year}{2014}\natexlab{}.
\newblock \showarticletitle{Microsoft COCO: Common objects in context}.
\newblock In \bibinfo{booktitle}{\emph{ECCV}}.
\newblock


\bibitem[\protect\citeauthoryear{Liu, Jin, Xu, Gong, and Mu}{Liu
  et~al\mbox{.}}{2020a}]%
        {vrd_with_st_global_context}
\bibfield{author}{\bibinfo{person}{Chenchen Liu}, \bibinfo{person}{Yang Jin},
  \bibinfo{person}{Kehan Xu}, \bibinfo{person}{Guoqiang Gong}, {and}
  \bibinfo{person}{Yadong Mu}.} \bibinfo{year}{2020}\natexlab{a}.
\newblock \showarticletitle{Beyond Short-Term Snippet: Video Relation Detection
  with Spatio-Temporal Global Context}. In \bibinfo{booktitle}{\emph{CVPR}}.
\newblock


\bibitem[\protect\citeauthoryear{Liu, Zhang, Chen, Wang, and Ouyang}{Liu
  et~al\mbox{.}}{2020b}]%
        {action_Liu_2020_CVPR}
\bibfield{author}{\bibinfo{person}{Ziyu Liu}, \bibinfo{person}{Hongwen Zhang},
  \bibinfo{person}{Zhenghao Chen}, \bibinfo{person}{Zhiyong Wang}, {and}
  \bibinfo{person}{Wanli Ouyang}.} \bibinfo{year}{2020}\natexlab{b}.
\newblock \showarticletitle{Disentangling and Unifying Graph Convolutions for
  Skeleton-Based Action Recognition}. In \bibinfo{booktitle}{\emph{CVPR}}.
\newblock


\bibitem[\protect\citeauthoryear{Mehta, Sotnychenko, Mueller, Xu, Elgharib,
  Fua, Seidel, Rhodin, Pons-Moll, and Theobalt}{Mehta et~al\mbox{.}}{2020}]%
        {XNect_SIGGRAPH2020}
\bibfield{author}{\bibinfo{person}{Dushyant Mehta}, \bibinfo{person}{Oleksandr
  Sotnychenko}, \bibinfo{person}{Franziska Mueller}, \bibinfo{person}{Weipeng
  Xu}, \bibinfo{person}{Mohamed Elgharib}, \bibinfo{person}{Pascal Fua},
  \bibinfo{person}{Hans-Peter Seidel}, \bibinfo{person}{Helge Rhodin},
  \bibinfo{person}{Gerard Pons-Moll}, {and} \bibinfo{person}{Christian
  Theobalt}.} \bibinfo{year}{2020}\natexlab{}.
\newblock \showarticletitle{{XNect}: Real-time Multi-Person {3D} Motion Capture
  with a Single {RGB} Camera}.
\newblock \bibinfo{journal}{\emph{ACM Transactions on Graphics}}.
\newblock


\bibitem[\protect\citeauthoryear{Mittal, Guhan, Bhattacharya, Chandra, Bera,
  and Manocha}{Mittal et~al\mbox{.}}{2020}]%
        {Mittal_2020_CVPR}
\bibfield{author}{\bibinfo{person}{Trisha Mittal}, \bibinfo{person}{Pooja
  Guhan}, \bibinfo{person}{Uttaran Bhattacharya}, \bibinfo{person}{Rohan
  Chandra}, \bibinfo{person}{Aniket Bera}, {and} \bibinfo{person}{Dinesh
  Manocha}.} \bibinfo{year}{2020}\natexlab{}.
\newblock \showarticletitle{EmotiCon: Context-Aware Multimodal Emotion
  Recognition Using Frege's Principle}. In \bibinfo{booktitle}{\emph{CVPR}}.
\newblock


\bibitem[\protect\citeauthoryear{Nagrani, Sun, Ross, Sukthankar, Schmid, and
  Zisserman}{Nagrani et~al\mbox{.}}{2020}]%
        {Nagrani_2020_CVPR}
\bibfield{author}{\bibinfo{person}{Arsha Nagrani}, \bibinfo{person}{Chen Sun},
  \bibinfo{person}{David Ross}, \bibinfo{person}{Rahul Sukthankar},
  \bibinfo{person}{Cordelia Schmid}, {and} \bibinfo{person}{Andrew Zisserman}.}
  \bibinfo{year}{2020}\natexlab{}.
\newblock \showarticletitle{Speech2Action: Cross-Modal Supervision for Action
  Recognition}. In \bibinfo{booktitle}{\emph{CVPR}}.
\newblock


\bibitem[\protect\citeauthoryear{Newell, Yang, and Deng}{Newell
  et~al\mbox{.}}{2016}]%
        {hourglass}
\bibfield{author}{\bibinfo{person}{Alejandro Newell}, \bibinfo{person}{Kaiyu
  Yang}, {and} \bibinfo{person}{Jia Deng}.} \bibinfo{year}{2016}\natexlab{}.
\newblock \showarticletitle{Stacked hourglass networks for human pose
  estimation}. In \bibinfo{booktitle}{\emph{ECCV}}.
\newblock


\bibitem[\protect\citeauthoryear{Omran, Lassner, Pons-Moll, Gehler, and
  Schiele}{Omran et~al\mbox{.}}{2018}]%
        {omran2018NBF}
\bibfield{author}{\bibinfo{person}{Mohamed Omran}, \bibinfo{person}{Christoph
  Lassner}, \bibinfo{person}{Gerard Pons-Moll}, \bibinfo{person}{Peter Gehler},
  {and} \bibinfo{person}{Bernt Schiele}.} \bibinfo{year}{2018}\natexlab{}.
\newblock \showarticletitle{Neural Body Fitting: Unifying Deep Learning and
  Model Based Human Pose and Shape Estimation}. In
  \bibinfo{booktitle}{\emph{3DV}}.
\newblock


\bibitem[\protect\citeauthoryear{Patel, Liao, and Pons-Moll}{Patel
  et~al\mbox{.}}{2020}]%
        {patel20tailornet}
\bibfield{author}{\bibinfo{person}{Chaitanya Patel},
  \bibinfo{person}{Zhouyingcheng Liao}, {and} \bibinfo{person}{Gerard
  Pons-Moll}.} \bibinfo{year}{2020}\natexlab{}.
\newblock \showarticletitle{TailorNet: Predicting Clothing in 3D as a Function
  of Human Pose, Shape and Garment Style}. In \bibinfo{booktitle}{\emph{CVPR}}.
\newblock


\bibitem[\protect\citeauthoryear{Pele and Werman}{Pele and Werman}{2008}]%
        {sift}
\bibfield{author}{\bibinfo{person}{O. Pele} {and} \bibinfo{person}{M. Werman}.}
  \bibinfo{year}{2008}\natexlab{}.
\newblock \showarticletitle{A linear time histogram metric for improved sift
  matching}. In \bibinfo{booktitle}{\emph{ECCV}}.
\newblock


\bibitem[\protect\citeauthoryear{Qi, Wang, Jia, Shen, and Zhu}{Qi
  et~al\mbox{.}}{2018}]%
        {GPNN}
\bibfield{author}{\bibinfo{person}{Siyuan Qi}, \bibinfo{person}{Wenguan Wang},
  \bibinfo{person}{Baoxiong Jia}, \bibinfo{person}{Jianbing Shen}, {and}
  \bibinfo{person}{Song-Chun Zhu}.} \bibinfo{year}{2018}\natexlab{}.
\newblock \showarticletitle{Learning Human-Object Interactions by Graph Parsing
  Neural Networks}. In \bibinfo{booktitle}{\emph{ECCV}}.
\newblock


\bibitem[\protect\citeauthoryear{Qian, Zhuang, Li, Xiao, Pu, and Xiao}{Qian
  et~al\mbox{.}}{2019}]%
        {vrd_st_graph}
\bibfield{author}{\bibinfo{person}{Xufeng Qian}, \bibinfo{person}{Yueting
  Zhuang}, \bibinfo{person}{Yimeng Li}, \bibinfo{person}{Shaoning Xiao},
  \bibinfo{person}{Shiliang Pu}, {and} \bibinfo{person}{Jun Xiao}.}
  \bibinfo{year}{2019}\natexlab{}.
\newblock \showarticletitle{Video Relation Detection with Spatio-Temporal
  Graph}. In \bibinfo{booktitle}{\emph{ACM MM}}.
\newblock


\bibitem[\protect\citeauthoryear{Randhavane, Bera, Kapsaskis, Sheth, Gray, and
  Manocha}{Randhavane et~al\mbox{.}}{2019}]%
        {EVA}
\bibfield{author}{\bibinfo{person}{Tanmay Randhavane}, \bibinfo{person}{Aniket
  Bera}, \bibinfo{person}{Kyra Kapsaskis}, \bibinfo{person}{Rahul Sheth},
  \bibinfo{person}{Kurt Gray}, {and} \bibinfo{person}{Dinesh Manocha}.}
  \bibinfo{year}{2019}\natexlab{}.
\newblock \showarticletitle{EVA: Generating Emotional Behavior of Virtual
  Agents Using Expressive Features of Gait and Gaze}. In
  \bibinfo{booktitle}{\emph{ACM Symposium on Applied Perception}}.
\newblock


\bibitem[\protect\citeauthoryear{Ren, He, Girshick, and Sun}{Ren
  et~al\mbox{.}}{2015}]%
        {NIPS2015_5638}
\bibfield{author}{\bibinfo{person}{Shaoqing Ren}, \bibinfo{person}{Kaiming He},
  \bibinfo{person}{Ross Girshick}, {and} \bibinfo{person}{Jian Sun}.}
  \bibinfo{year}{2015}\natexlab{}.
\newblock \showarticletitle{Faster R-CNN: Towards Real-Time Object Detection
  with Region Proposal Networks}.
\newblock In \bibinfo{booktitle}{\emph{NeurIPS}}.
\newblock


\bibitem[\protect\citeauthoryear{Shang, Di, Xiao, Cao, Yang, and Chua}{Shang
  et~al\mbox{.}}{2019}]%
        {annotating_objects_relations}
\bibfield{author}{\bibinfo{person}{Xindi Shang}, \bibinfo{person}{Donglin Di},
  \bibinfo{person}{Junbin Xiao}, \bibinfo{person}{Yu Cao}, \bibinfo{person}{Xun
  Yang}, {and} \bibinfo{person}{Tat-Seng Chua}.}
  \bibinfo{year}{2019}\natexlab{}.
\newblock \showarticletitle{Annotating Objects and Relations in User-Generated
  Videos}. In \bibinfo{booktitle}{\emph{ICMR}}.
\newblock


\bibitem[\protect\citeauthoryear{Shang, Ren, Guo, Zhang, and Chua}{Shang
  et~al\mbox{.}}{2017}]%
        {video_vrd}
\bibfield{author}{\bibinfo{person}{Xindi Shang}, \bibinfo{person}{Tongwei Ren},
  \bibinfo{person}{Jingfan Guo}, \bibinfo{person}{Hanwang Zhang}, {and}
  \bibinfo{person}{Tat-Seng Chua}.} \bibinfo{year}{2017}\natexlab{}.
\newblock \showarticletitle{Video Visual Relation Detection}. In
  \bibinfo{booktitle}{\emph{ACM MM}}.
\newblock


\bibitem[\protect\citeauthoryear{Shi, Zhang, Cheng, and Lu}{Shi
  et~al\mbox{.}}{2019}]%
        {2sagcn2019cvpr}
\bibfield{author}{\bibinfo{person}{Lei Shi}, \bibinfo{person}{Yifan Zhang},
  \bibinfo{person}{Jian Cheng}, {and} \bibinfo{person}{Hanqing Lu}.}
  \bibinfo{year}{2019}\natexlab{}.
\newblock \showarticletitle{Two-Stream Adaptive Graph Convolutional Networks
  for Skeleton-Based Action Recognition}. In \bibinfo{booktitle}{\emph{CVPR}}.
\newblock


\bibitem[\protect\citeauthoryear{Sun, Ren, Zi, and Wu}{Sun
  et~al\mbox{.}}{2019}]%
        {vrd_feature_fusion}
\bibfield{author}{\bibinfo{person}{Xu Sun}, \bibinfo{person}{Tongwei Ren},
  \bibinfo{person}{Yuan Zi}, {and} \bibinfo{person}{Gangshan Wu}.}
  \bibinfo{year}{2019}\natexlab{}.
\newblock \showarticletitle{Video Visual Relation Detection via Multi-modal
  Feature Fusion}. In \bibinfo{booktitle}{\emph{ACM MM}}.
\newblock


\bibitem[\protect\citeauthoryear{Tsai, Divvala, Morency, Salakhutdinov, and
  Farhadi}{Tsai et~al\mbox{.}}{2019}]%
        {gated_st_energy_graph}
\bibfield{author}{\bibinfo{person}{Yao-Hung~Hubert Tsai},
  \bibinfo{person}{Santosh Divvala}, \bibinfo{person}{Louis-Philippe Morency},
  \bibinfo{person}{Ruslan Salakhutdinov}, {and} \bibinfo{person}{Ali Farhadi}.}
  \bibinfo{year}{2019}\natexlab{}.
\newblock \showarticletitle{Video Relationship Reasoning using Gated
  Spatio-Temporal Energy Graph}. In \bibinfo{booktitle}{\emph{CVPR}}.
\newblock


\bibitem[\protect\citeauthoryear{Varol, Romero, Martin, Mahmood, Black, Laptev,
  and Schmid}{Varol et~al\mbox{.}}{2017}]%
        {Varol_2017_CVPR}
\bibfield{author}{\bibinfo{person}{Gul Varol}, \bibinfo{person}{Javier Romero},
  \bibinfo{person}{Xavier Martin}, \bibinfo{person}{Naureen Mahmood},
  \bibinfo{person}{Michael~J. Black}, \bibinfo{person}{Ivan Laptev}, {and}
  \bibinfo{person}{Cordelia Schmid}.} \bibinfo{year}{2017}\natexlab{}.
\newblock \showarticletitle{Learning From Synthetic Humans}. In
  \bibinfo{booktitle}{\emph{CVPR}}.
\newblock


\bibitem[\protect\citeauthoryear{Wan, Zhou, Liu, Li, and He}{Wan
  et~al\mbox{.}}{2019}]%
        {PMFNet}
\bibfield{author}{\bibinfo{person}{Bo Wan}, \bibinfo{person}{Desen Zhou},
  \bibinfo{person}{Yongfei Liu}, \bibinfo{person}{Rongjie Li}, {and}
  \bibinfo{person}{Xuming He}.} \bibinfo{year}{2019}\natexlab{}.
\newblock \showarticletitle{Pose-aware Multi-level Feature Network for Human
  Object Interaction Detection}. In \bibinfo{booktitle}{\emph{ICCV}}.
\newblock


\bibitem[\protect\citeauthoryear{Wang, Xiao, Jiang, Shao, Sun, and Shen}{Wang
  et~al\mbox{.}}{2018}]%
        {reploss}
\bibfield{author}{\bibinfo{person}{Xinlong Wang}, \bibinfo{person}{Tete Xiao},
  \bibinfo{person}{Yuning Jiang}, \bibinfo{person}{Shuai Shao},
  \bibinfo{person}{Jian Sun}, {and} \bibinfo{person}{Chunhua Shen}.}
  \bibinfo{year}{2018}\natexlab{}.
\newblock \showarticletitle{Repulsion Loss: Detecting Pedestrians in a Crowd}.
  In \bibinfo{booktitle}{\emph{CVPR}}.
\newblock


\bibitem[\protect\citeauthoryear{Xiong, Kim, and Singh}{Xiong
  et~al\mbox{.}}{2019}]%
        {Xiong_2019_CVPR}
\bibfield{author}{\bibinfo{person}{Yunyang Xiong}, \bibinfo{person}{Hyunwoo~J.
  Kim}, {and} \bibinfo{person}{Vikas Singh}.} \bibinfo{year}{2019}\natexlab{}.
\newblock \showarticletitle{Mixed Effects Neural Networks (MeNets) With
  Applications to Gaze Estimation}. In \bibinfo{booktitle}{\emph{CVPR}}.
\newblock


\bibitem[\protect\citeauthoryear{Xu, Li, Wong, Kankanhalli, and Zhao}{Xu
  et~al\mbox{.}}{2018}]%
        {intend}
\bibfield{author}{\bibinfo{person}{Bingjie Xu}, \bibinfo{person}{Junnan Li},
  \bibinfo{person}{Yongkang Wong}, \bibinfo{person}{Mohan~S Kankanhalli}, {and}
  \bibinfo{person}{Qi Zhao}.} \bibinfo{year}{2018}\natexlab{}.
\newblock \showarticletitle{Interact as you intend: Intention driven
  human-object interaction detection}. In \bibinfo{booktitle}{\emph{arXiv
  preprint arXiv:1808.09796}}.
\newblock


\bibitem[\protect\citeauthoryear{Xu, Wong, Li, Zhao, and Kankanhalli}{Xu
  et~al\mbox{.}}{2019}]%
        {knowledgeGraph}
\bibfield{author}{\bibinfo{person}{B. Xu}, \bibinfo{person}{Y. Wong},
  \bibinfo{person}{J. Li}, \bibinfo{person}{Q. Zhao}, {and}
  \bibinfo{person}{M.~S. Kankanhalli}.} \bibinfo{year}{2019}\natexlab{}.
\newblock \showarticletitle{Learning to Detect Human-Object Interactions With
  Knowledge}. In \bibinfo{booktitle}{\emph{CVPR}}.
\newblock


\bibitem[\protect\citeauthoryear{Yao and Fei-Fei}{Yao and Fei-Fei}{2010}]%
        {grouplet}
\bibfield{author}{\bibinfo{person}{B. Yao} {and} \bibinfo{person}{L. Fei-Fei}.}
  \bibinfo{year}{2010}\natexlab{}.
\newblock \showarticletitle{Grouplet: A structured image representation for
  recognizing human and object interactions.}. In
  \bibinfo{booktitle}{\emph{CVPR}}.
\newblock


\bibitem[\protect\citeauthoryear{Yao, Jiang, Khosla, Lin, Guibas, and
  Fei-Fei}{Yao et~al\mbox{.}}{2011}]%
        {actionbases}
\bibfield{author}{\bibinfo{person}{Bangpeng Yao}, \bibinfo{person}{Xiaoye
  Jiang}, \bibinfo{person}{Aditya Khosla}, \bibinfo{person}{Andy~Lai Lin},
  \bibinfo{person}{Leonidas Guibas}, {and} \bibinfo{person}{Li Fei-Fei}.}
  \bibinfo{year}{2011}\natexlab{}.
\newblock \showarticletitle{Human Action Recognition by Learning Bases of
  Action Attributes and Parts}. In \bibinfo{booktitle}{\emph{ICCV}}.
\newblock


\bibitem[\protect\citeauthoryear{Zhang, Lan, Zeng, Xing, Xue, and Zheng}{Zhang
  et~al\mbox{.}}{2020}]%
        {action_Zhang_2020_CVPR}
\bibfield{author}{\bibinfo{person}{Pengfei Zhang}, \bibinfo{person}{Cuiling
  Lan}, \bibinfo{person}{Wenjun Zeng}, \bibinfo{person}{Junliang Xing},
  \bibinfo{person}{Jianru Xue}, {and} \bibinfo{person}{Nanning Zheng}.}
  \bibinfo{year}{2020}\natexlab{}.
\newblock \showarticletitle{Semantics-Guided Neural Networks for Efficient
  Skeleton-Based Human Action Recognition}. In
  \bibinfo{booktitle}{\emph{CVPR}}.
\newblock


\bibitem[\protect\citeauthoryear{Zhang, Wen, Bian, Lei, and Li}{Zhang
  et~al\mbox{.}}{2018}]%
        {Zhang_2018_ECCV}
\bibfield{author}{\bibinfo{person}{Shifeng Zhang}, \bibinfo{person}{Longyin
  Wen}, \bibinfo{person}{Xiao Bian}, \bibinfo{person}{Zhen Lei}, {and}
  \bibinfo{person}{Stan~Z. Li}.} \bibinfo{year}{2018}\natexlab{}.
\newblock \showarticletitle{Occlusion-aware R-CNN: Detecting Pedestrians in a
  Crowd}. In \bibinfo{booktitle}{\emph{ECCV}}.
\newblock


\end{thebibliography}

\end{document}